\documentclass[sigconf]{acmart}

\AtBeginDocument{%
  \providecommand\BibTeX{{%
    \normalfont B\kern-0.5em{\scshape i\kern-0.25em b}\kern-0.8em\TeX}}}

\setcopyright{acmlicensed}
\copyrightyear{2024}
\acmYear{2024}
\acmDOI{XXXXXXX.XXXXXXX}

\acmConference[RecSys 2024]{Make sure to enter the correct
  conference title from your rights confirmation email}{October 14–18, 2024}{Bari, Italy}
\acmISBN{978-1-4503-XXXX-X/18/06}

\graphicspath{{./images/}}

\begin{document}

\title{Integrating Domain Knowledge into Large Language Models for Enhanced
Fashion Recommendations}

\author{Zhan Shi}
\authornote{Both authors contributed equally to this research.}
\email{aria2@scu.edu}
\authornotemark[1]
\affiliation{%
  \institution{Santa Clara University}
  \streetaddress{}
  \city{Santa Clara}
  \country{USA}
}

\author{Shanglin Yang}
\authornote{Both authors contributed equally to this research.}
\affiliation{%
  \institution{}
  \streetaddress{}
  \city{}
  \country{}}
\email{kudoysl@gmail.com}


\begin{abstract}

Fashion, deeply rooted in sociocultural dynamics, evolves as individuals emulate styles popularized by influencers and iconic figures. In the quest to replicate such refined tastes using artificial intelligence, traditional fashion ensemble methods have primarily used supervised learning to imitate the decisions of style icons, which falter when faced with distribution shifts, leading to style replication discrepancies triggered by slight variations in input. Meanwhile, large language models (LLMs) have become prominent across various sectors, recognized for their user-friendly interfaces, strong conversational skills, and advanced reasoning capabilities. To address these challenges, we introduce the Fashion Large Language Model (FLLM), which employs auto-prompt generation training strategies to enhance its capacity for delivering personalized fashion advice while retaining essential domain knowledge. Additionally, by integrating a retrieval augmentation technique during inference, the model can better adjust to individual preferences. Our results show that this approach surpasses existing models in accuracy, interpretability, and few-shot learning capabilities.
\end{abstract}



\maketitle

\section{Introduction}

Fashion plays a significant role in various aspects of life, including social and cultural identity. People often imitate styles suggested by fashion experts, icons, or Key Opinion Leaders (KOLs) from social media. In the rapidly evolving fashion technology sector, creating advanced, personalized recommendation systems is a major academic and commercial endeavor. This research introduces an innovative approach using Large Language Models (LLMs) to transform fashion recommendations. Unlike traditional methods relying on static datasets and supervised learning, our generative model leverages LLMs to provide personalized, flexible fashion advice. This approach overcomes previous limitations and offers a deeper connection with individual preferences, reflecting the dynamic nature of fashion.
\begin{figure}
\centering
\includegraphics[width=1.00 \linewidth]{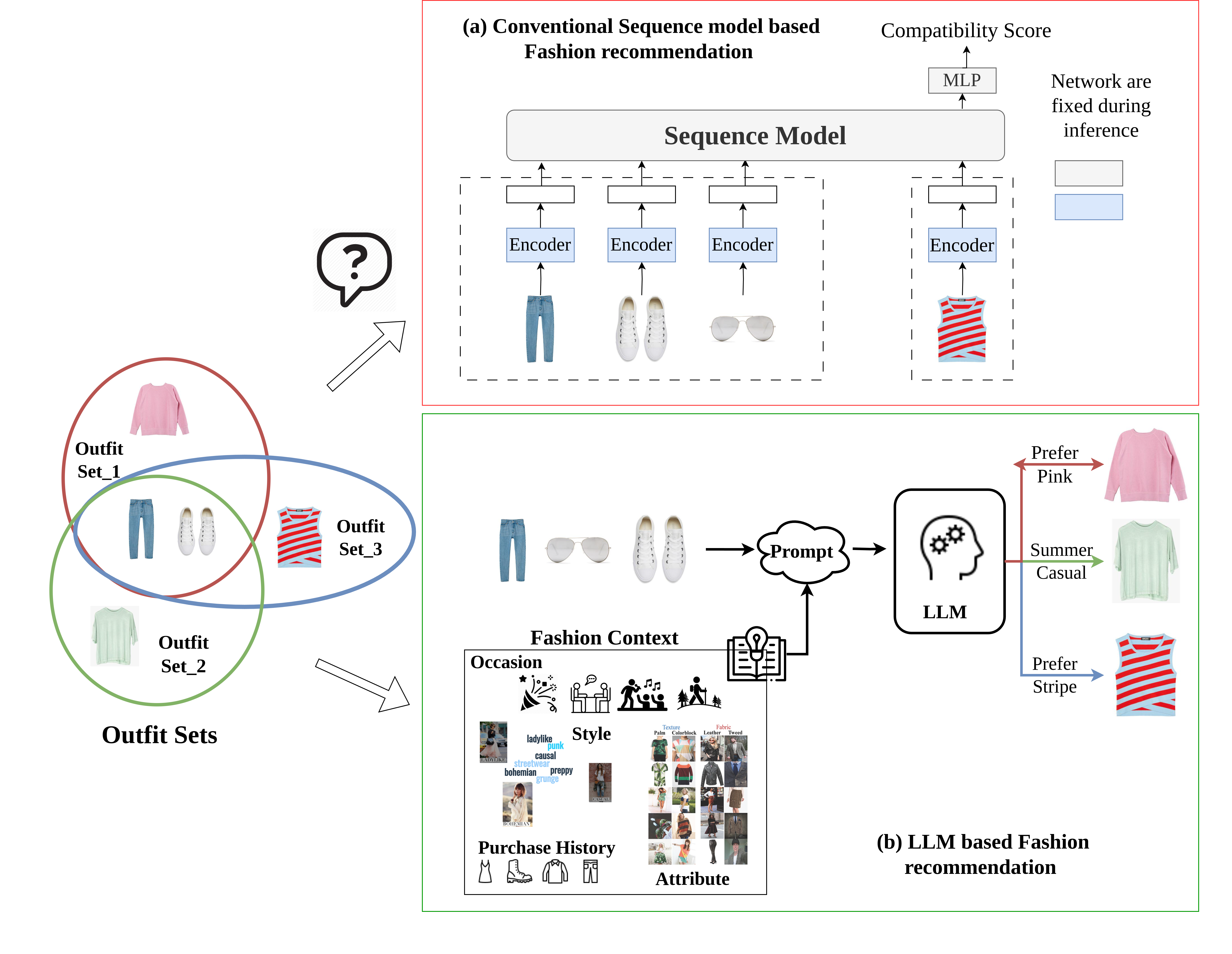}
\caption{Fashion item recommendations can differ significantly based on various occasions, styles, and attributes. (a) Traditional methods (red box), trained solely on fixed datasets using sequence models, are constrained to aligning with the training data and struggle to adapt to new preference distributions. In contrast, our (b) LLM methods (green box) dynamically integrate diverse contexts into the prompt. Leveraging the inference capabilities of large models, our methods produce results under a wide range of conditions.}
\label{fig:outfit_sample}
\end{figure}
Fashion trends change rapidly, posing challenges for recommendation systems to forecast and adapt to personal preferences. Conventional systems, though somewhat successful in predicting item compatibility and ensuring stylistic consistency, often fail to cater to diverse individual styles and rely heavily on static datasets, limiting their relevance over time.

To address these challenges, our study presents an innovative solution using LLMs' generative and adaptive strengths \cite{wu2023brief}, offering new methods for generating recommendations that include items, occasions, and styles. We propose a new training paradigm for LLMs, starting with fashion-specific data to retain core knowledge and progressing through specialized training phases. This includes a foundational phase with 'FIIB' data, a combined phase of style and 'FIIB' data, and the development of a custom dataset for final refinement.

The paper is structured as follows: We first explore the current landscape of fashion recommendation systems, highlighting their limitations and the potential of LLMs. We then detail our novel approach, including the architecture of our LLM-based system, data sources, training procedure, and the use of retrieval-augmented generation. Following this, we describe our experimental design, evaluation metrics, and empirical study results, demonstrating our model's superior performance in accuracy, adaptability, and user satisfaction. We conclude with a discussion of the broader implications and future research directions.
Our key contributions are:
\begin{itemize}
  \item Reconceptualizing fashion recommendation as a generative challenge using LLMs.
  \item Incorporating domain fine-tuning and retrieval-augmented generation to enable the model to learn new fashion knowledge while preserving domain expertise.
  \item Demonstrating the model's performance through experiments, highlighting improved accuracy, impressive few-shot learning capabilities, and adaptability to different contexts.
\end{itemize}

\section{RELATED WORK}
In recent years, the use of deep learning neural networks, highlighted by \cite{10.1145/3447239}, has drawn significant attention in various tasks within the fashion industry. These tasks include category and attribute classification (\cite{papadopoulos2022multimodal}), trend forecasting (\cite{al2017fashion}, popularity prediction (\cite{papadopoulos2022victor}), and the development of fashion recommendation systems (\cite{hwangbo2018recommendation}, \cite{stefani2019cfrs}), particularly in outfit recommendation. It's crucial to determine which garments go well together to recommend complete outfits that are compatible and cohesive.

Initial attempts to address outfit compatibility treated it as a series of pairwise comparisons among all garments in an outfit, as explored by \cite{Vasileva_2018_ECCV}. These pairwise-based methods used Siamese networks (\cite{veit2015learning}) and triplet loss networks, with either type-aware embeddings (\cite{vasileva2018learning}) or similarity-aware embeddings (\cite{vasileva2018learning}). In contrast, some researchers have aimed to capture a holistic view of outfit-level representations through bidirectional LSTMs (\cite{han2017learning}) or graph neural networks (\cite{ding2023computational}, \cite{liu2020learning}). Some studies also have shifted to attention-based methods, including the Transformer architecture for personalized outfit recommendations and complementary item retrieval (\cite{mo2023towards}, \cite{xie2021personalized}, \cite{sarkar2023outfittransformer}).

Researchers have increasingly turned to Large Language Models (LLMs) for their outstanding text generation capabilities, particularly for data augmentation (\cite{ke2023critiquellm}, \cite{liu2024once}). \cite{carbune2024chart}, for instance, used LLMs to create multimodal datasets that blend language and images for instruction-following tasks. Their method, which involved tuning the instructions based on this data, significantly improved both vision and language comprehension.

In the domain of personalized recommendation systems, LLMs have also played a crucial role. \cite{yang2023palr} employed LLMs to generate user profiles by incorporating behaviors like clicks, purchases, and ratings. These profiles, when combined with the history of user interactions and potential items, were key in formulating the final recommendation prompt. LLMs were then applied to determine the likelihood of user-item interactions from this prompt. In a similar vein, \cite{lyu2023llm} developed a technique that leverages LLMs to merge reasoning about user preferences with factual knowledge of items. However, our research explores a new area: there is no established work on using LLMs in fashion recommendation. Therefore, our study establishes a novel baseline for LLM application in the fashion recommendation field.

\begin{figure*}
\begin{center}
    \includegraphics[width=1.00\linewidth]{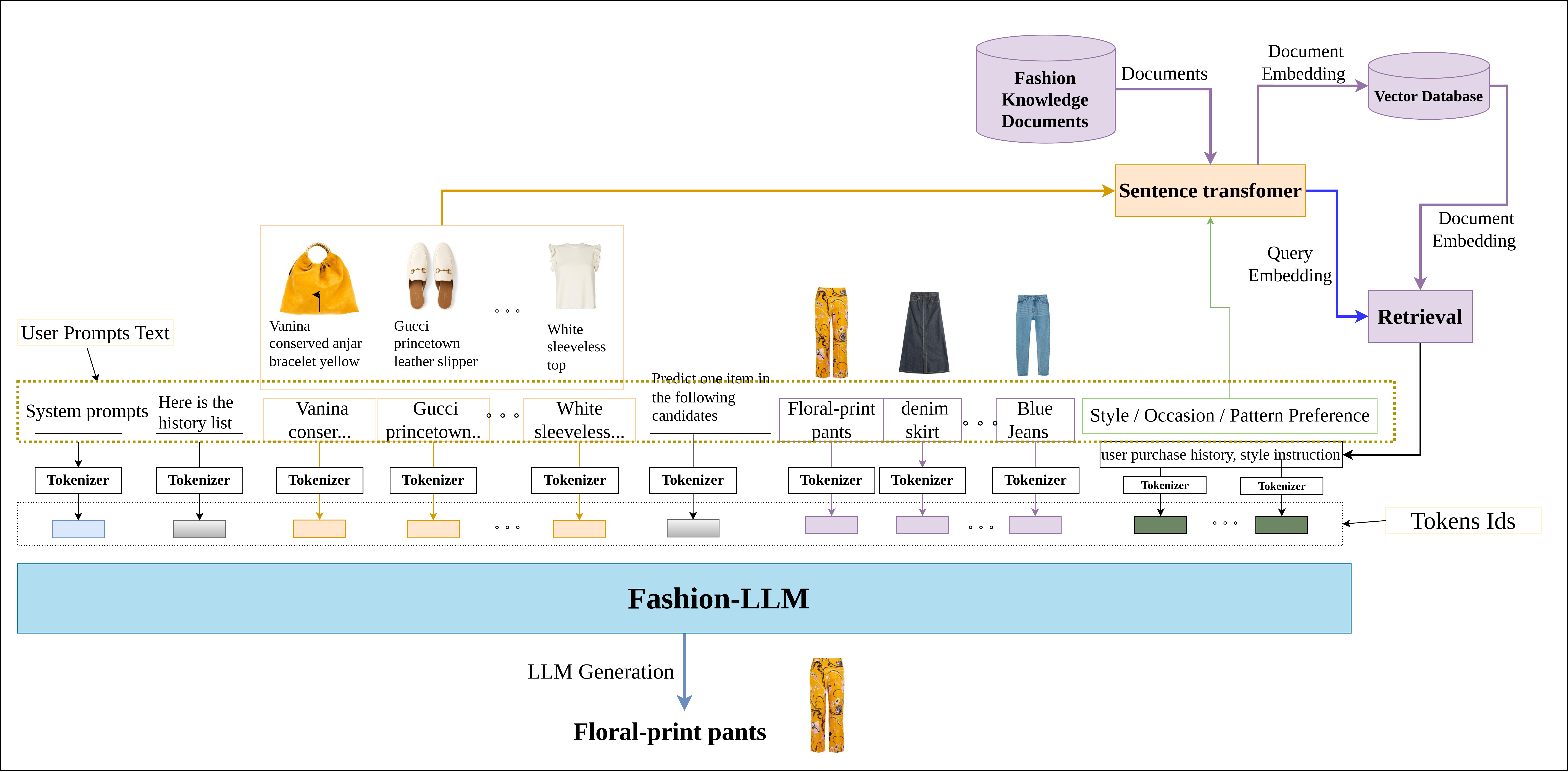}
\end{center}
\caption{An overview of an information-augmented retrieval system featuring a fine-tuned Large Language Model (Fashion-LLM) is presented. This system integrates multiple inputs into the Fashion-LLM to generate precise product descriptions.}
    \label{fig:fashionllm_sys}
\end{figure*}

\section{Methodology}
\subsection{Problem Formulation}
Figure~\ref{fig:fashionllm_sys} illustrates our sophisticated fashion recommendation system, which utilizes a finely-tuned Large Language Model (LLM) integrated with domain-specific retrieval processes. This system, optimized with a dedicated fashion dataset, is adept at parsing the complexities of fashion nuances. It employs the LLM to create specific prompts that trigger searches within a comprehensive fashion product dataset, and a dual-component mechanism retrieves vital domain context by merging industry trends and design principles with user preferences to tailor recommendations.

The system processes both user and system-generated text via tokenizers, converting them into tokens suitable for the LLM. The model then analyzes these inputs to generate customized recommendations, such as suggesting "floral-print pants" among various item choices. Fashion-related documents are processed into embeddings through a sentence transformer and stored in a vector database. These embeddings are then matched with user query embeddings to further refine the recommendations.

Ultimately, the LLM delivers sophisticated responses to user queries, resulting in either well-coordinated outfit recommendations or detailed fashion advice. This ensures both contextual relevance and personalization. By integrating system prompts, item description and processed contextual data, the system achieves precise and personalized fashion recommendations.
\subsection{Fintuning LLM with auto-prompts generation}
\begin{figure}
\centering
    \includegraphics[width=1.00 \linewidth]{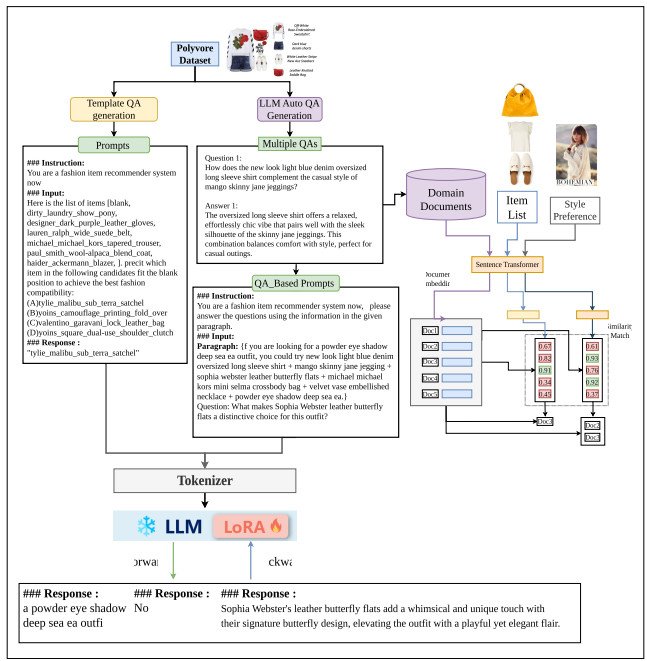}
  \caption{Detailed Workflow of Model Fine-Tuning and Retrieval Processes for Fashion Recommender System. In this AI-powered system, we employ both template-based AQ generation and LLM auto-question generation to create domain-relevant questions and prompts. }
    \label{fig:llm_finetuning}
\end{figure}
The diagram illustrated in Figure \ref{fig:llm_finetuning} delineates the comprehensive training and refinement process for a specialized Fashion Large Language Model (FLLM), which is tasked with dispensing tailored fashion advice. Initially, the FLLM is acquainted with Polyvore outfit training data. This data encompasses two primary types: binary question data, which prompts the model to discern the appropriateness of an outfit, and Fill-in-the-Blank (FITB) data.

To enhance the robustness and proficiency of our Fashion Large Language Model (FLLM), we employ a dual-strategy approach in generating training data. Initially, we source training data directly from existing datasets. Additionally, we utilize the advanced capabilities of a domain-specific model to further strengthen the FLLM.
\vspace{-\baselineskip}
\begin{enumerate}
\item \textbf{Template QA Generation}: This data comprises a list of items for which the model generates a binary response—assessing whether the items constitute a compatible outfit and suggesting potential item candidates as suitable choices.

\item \textbf{LLM Auto QA Generation}: The training prompts are crafted to elicit detailed descriptions that highlight the style and fit of outfits. This enriches the model with comprehensive textual content, preserving and utilizing domain knowledge to refine its general understanding. Initially, the model generates multiple QAs based on fashion style, which are then converted into template training data. This also supports the domain knowledge needed for the retrieval module.
\end{enumerate}

These strategies, as reflected in the designed prompts, ensure that the FLLM not only retains its core language processing capabilities but is also finely tuned to analyze and address fashion-related queries with enhanced precision. Leveraging insights from domain knowledge, including broad fashion trends and specific user purchasing behaviors, the model crafts tailored training prompts. These prompts are instrumental in boosting the FLLM's predictive accuracy and its capacity to deliver customized fashion recommendations.
By diversifying the retrieval process through these multi-path queries, the model significantly broadens its ability to gather information from various facets, culminating in a richly layered final context. This multi-aspect query enhancement allows the model not only to retrieve more relevant information but also to provide more nuanced and comprehensive fashion item recommendations.

\subsection{Retrieval augmented inference}
We introduce a novel architecture that incorporates a Retrieval-Augmented Generation (RAG) model within a Large Language Model (LLM) to enhance fashion item recommendations, as shown in Figure \ref{fig:fashionllm_sys}. This architecture improves upon traditional retrieval methods by dividing the retrieval process into multiple query pathways, each tailored to increase the contextual depth of the information retrieved.

The architecture employs direct embedding-based queries, utilizing the capabilities of neural embeddings for immediate and accurate retrieval of fashion items. It also incorporates queries tailored to style and occasion, enabling contextually aware recommendations that match specific user scenarios and stylistic preferences. The queries are aligned with the knowledge documented in our generated knowledge path, as outlined in Figure \ref{fig:llm_finetuning}. Additionally, it leverages queries derived from LLM-generated questions. This innovative approach generates dynamic questions that reflect current fashion trends and individual user profiles, further refining the retrieval process.

\section{Experiments}
\label{sec:eval}

\subsection{Dataset}
The Polyvore dataset \cite{han2017learning} is a robust, community-generated collection of fashion ensembles, each meticulously curated to ensure that the individual items within them are complementary. It encompasses an extensive compilation of 68,000 manually curated outfits. A specialized subset known as Polyvore-disjoint is derived from the primary Polyvore dataset. This subset is crafted by meticulously filtering out any outfits that share items across the training, validation, and testing sets, thereby ensuring a strict separation of data. 

\subsection{Evaluation}

\textbf{FITB accuracy}.
 We carried out a fill-in-the-blank accuracy assessment using the Polyvore dataset, as detailed in Table 1. The results indicate that, compared to traditional methods, the LLM demonstrates superior proficiency in grasping more complex contexts and achieves the highest accuracy in both Joint and Disjoint Dataset.
    
\begin{table}[ht]
\centering
\caption{Performance comparison of different methods on Polyvore Outfits-D and Polyvore Outfits datasets.}
\begin{tabular}{lcc}
\toprule
Method & \multicolumn{1}{c}{Polyvore Outfits-D} & \multicolumn{1}{c}{Polyvore Outfits} \\
\midrule
Type-Aware \cite{vasileva2018learning} & 55.65 & 57.83 \\
SCE-Net Average \cite{xing2022sce} & 53.67 & 59.07 \\
CSA-Net \cite{Lin_2020} & 59.26 & 63.73 \\
OutfitTransformer \cite{sarkar2023outfittransformer}  & 59.48 & 67.10 \\
FashionLLM(ours)  & \textbf{62.17} & \textbf{67.21} \\
\bottomrule
\end{tabular}
\end{table}

\begin{figure}[t]
\begin{center}
   \includegraphics[scale=0.3]{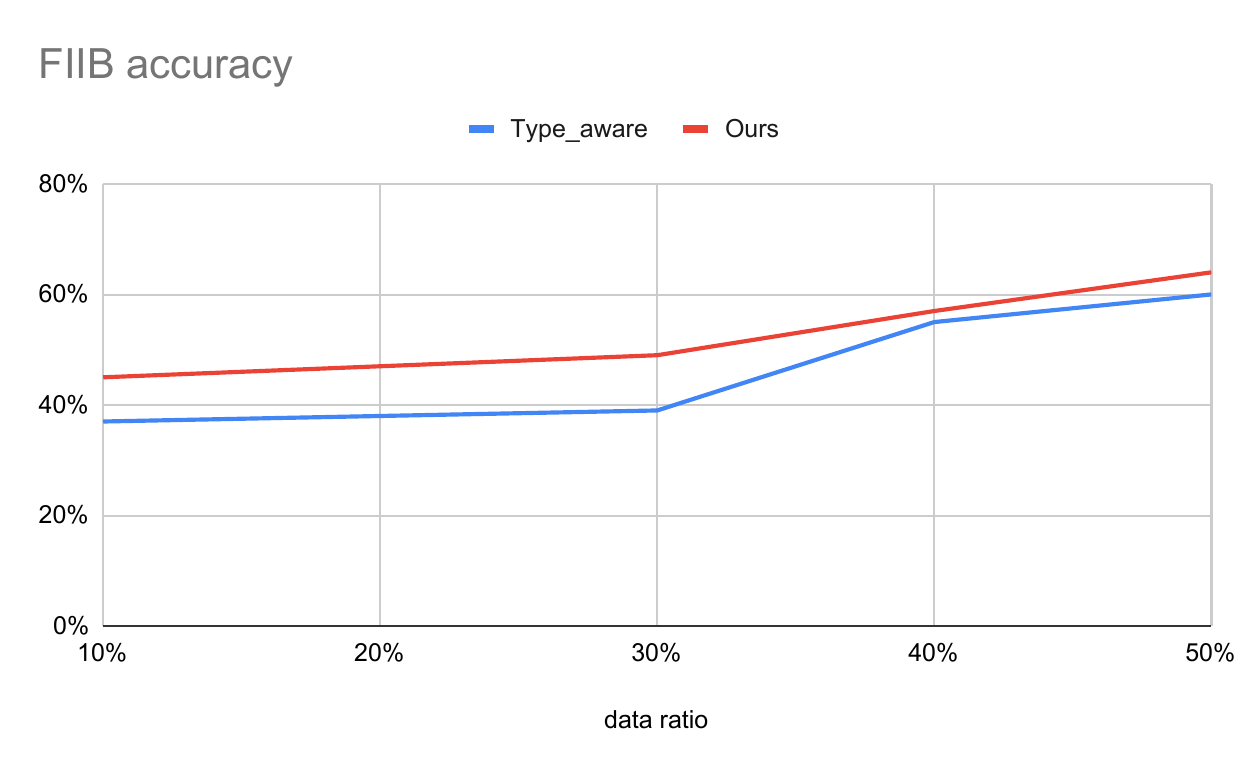}
\end{center}
   \caption{Performance of different methods with respect to FIIB at different  of ratio of training data}
\label{fig:map}
\end{figure}

\textbf{Few-shot ability}.
We also evaluated the performance of our model across different training data ratios to assess the impact of data volume on its effectiveness. A lower ratio indicates reduced data use for training. This evaluation compared our method to the state-of-the-art Type-Aware methods in fashion prediction. Our results show that although the final performance of both algorithms is similar when using the full dataset, a significant difference emerges at lower data ratios as shown in Figure \ref{fig:map}. Here, our algorithm surpasses the baseline by about 10 percent in accuracy improvement. This demonstrates the Large Language Model's (LLM) ability to effectively learn from limited data and maintain high prediction accuracy, which is especially advantageous in situations where gathering extensive datasets is challenging, providing a significant advantage in data-sparse environments.

\begin{figure}
    \centering
    \includegraphics[width=1.0\linewidth]{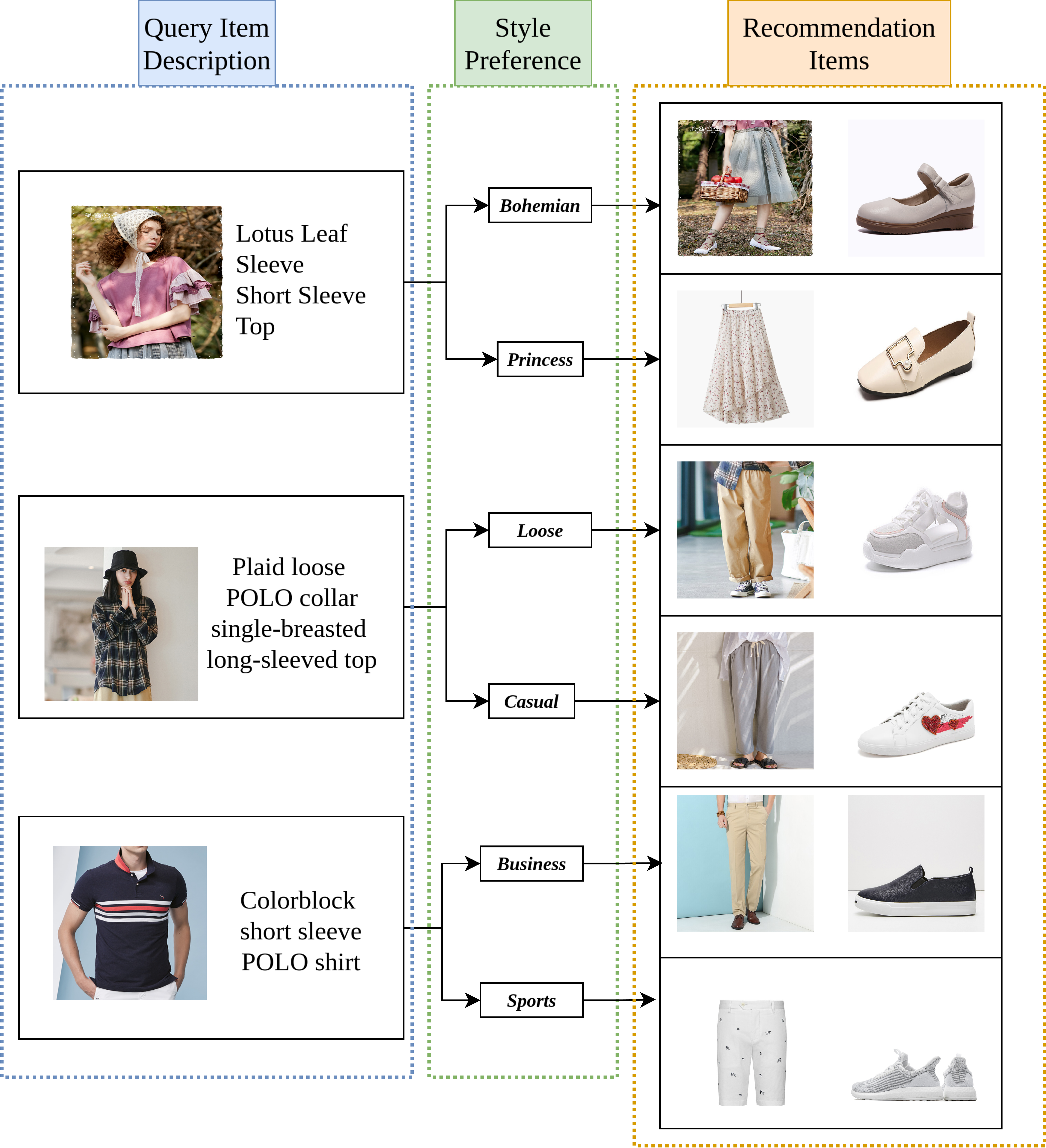}
    \caption{Visualization of recommendation results: For each query item, our model can generate different outfits based on various style preferences. }
    \label{fig:fashion_vis}
\end{figure}

\textbf{Visualizing Recommendation Results}. 
The visualization of recommendation results, as depicted in Figure \ref{fig:fashion_vis}, demonstrates how our approach can achieve diverse matching outcomes compatible with both the item and the style preferences. It effectively maps query items to style preferences and subsequently to corresponding recommendation items, showcasing the model's adaptability and effectiveness in aligning fashion choices with user preferences.

\section{Remarks and Future Work}
\label{sec:conclusion}

Our model has established a Large Language Model (LLM)-based fashion recommendation system characterized by high accuracy and dynamic adaptation. The LLM model can quickly adjust to user preferences and new data distributions. Moving forward, our focus will be on a hybrid approach that combines the outputs of the LLM and a Conventional Recommendation Model (CRM) to form a model ensemble. This ensemble will also incorporate multimodal information, including images and metadata to further develop the system's ability to process and integrate multimodal data, improving the accuracy and personalization of fashion recommendations.

\begin{acks}
Acknowledgements go here. Delete enclosing begin/end markers if there are no acknowledgements.
\end{acks}

\bibliographystyle{ACM-Reference-Format}
\bibliography{main.bib}

\end{document}